%% file: 25_huayi_ral.tex
\PassOptionsToPackage{pdftex, xetex}{graphicx}
\PassOptionsToPackage{usenames,dvipsnames,svgnames,table}{xcolor}
\documentclass[letterpaper, 10pt, conference]{ieeeconf}
\IEEEoverridecommandlockouts
\makeatletter
\let\NAT@parse\undefined
\makeatother
\usepackage[sort,numbers,compress]{natbib}
\makeatother

\usepackage{pratik}

\usepackage{makecell}

\graphicspath{{./figures/}}
\newlength{\mylen}
\makeatletter
\DeclareMathSizes{9.95}{9}{7}{5}
\DeclareMathSizes{10}{9}{7}{5}
\DeclareMathSizes{11}{10}{8}{6}
\makeatother

\title{\LARGE \bf
Active Semantic Perception
}

\author{
Huayi Tang and Pratik Chaudhari
\thanks{
The authors are with the
General Robotics, Automation, Sensing and Perception (GRASP) Laboratory at the University of Pennsylvania.\\
Email: \href{mailto:huayit@upenn.edu}{huayit@upenn.edu},
\href{mailto:pratikac@upenn.edu}{pratikac@upenn.edu}%
}
}

\begin{document}

\maketitle
\thispagestyle{empty}
\pagestyle{empty}

\input{./1-abstract}

\input{./2-introduction}

\input{./3-related_works}

\input{./4-method}

\input{./5-experiments}

\input{./6-future_work}

\input{./7-appendix}

\section*{ACKNOWLEDGMENT}
This work was supported by grants from the National Science Foundation \& USDA NIFA 2022-67021-36856 (NRI 3.0) and CMMI 2415249. We would also like to acknowledge Dimitris Dimos for his assistance in the experiments on the Unitree Go 2 robot.

{
\footnotesize
\setlength{\bibsep}{0ex}
\bibliographystyle{IEEEtran}
\bibliography{references, pratik}
}
\end{document}

%% file: 1-abstract.tex

\begin{abstract}

We develop an approach for active semantic perception, which refers to using the semantics of the scene for tasks such as exploration.
We build a compact, multi-layer scene graph that can represent large, complex indoor environments at various levels of abstraction, e.g., nodes corresponding to rooms, objects, walls, windows etc., as well as fine-grained details of their geometry.
We develop a procedure based on large language models (LLMs) to sample new plausible scene graphs of unobserved regions that are consistent with partial observations of the scene.
We develop a procedure to compute the information gain of a potential waypoint upon this scene graph to enable sophisticated spatial reasoning: for example, of the two doors that lead out of the living room, one probably leads to the kitchen and the other to the bedroom.
We evaluate our approach in realistic 3D indoor apartments in simulation and also on a Unitree Go 2 robot in the real world.
Qualitative and quantitative analysis shows that our approach can pin down high-level and low-level semantic information in the environment quickly and more accurately than existing approaches.
The source code and additional information can be found on \href{https://github.com/grasp-lyrl/active_semantic_perception}{https://github.com/grasp-lyrl/active\_semantic\_perception}.

\end{abstract}

%% file: 2-introduction.tex

\section{Introduction}

Suppose a firefighter is trying to rescue a person from a house.
This task requires understanding the environment, e.g., the locations of staircases, the arrangement of rooms, and the objects they contain.
Doing the task expeditiously also requires guessing as to where the person could be located.
Humans are spectacular at these tasks.
Robots cannot really perform these kinds of tasks yet.
This is not because they lack the ability to sense or to move.
Robot sensing and mobility are very good---if not as good as those of humans.

Perhaps the key distinguishing factor is that humans and robots build very different representations of space.
Robots are good at identifying geometric and photometric details of the scene or identifying objects in the scene---these are all syntactical entities. Human representations are much richer in their semantics---the relationships between these entities.
Our firefighters would find it strange if the apartment above had a bathtub inside the kitchen. Semantics is about us (humans) attributing meaning to objects in the scene \cite{he2025mathematical, kirby2019invitation,floyd1993assigning}.%
\footnote{In mathematical logic, semantics arises from interpreting syntax in mathematical structures \cite{kirby2019invitation}.}

\begin{figure*}[!htpb]
\vspace{4pt}
\centering
\includegraphics[width=0.90\linewidth]{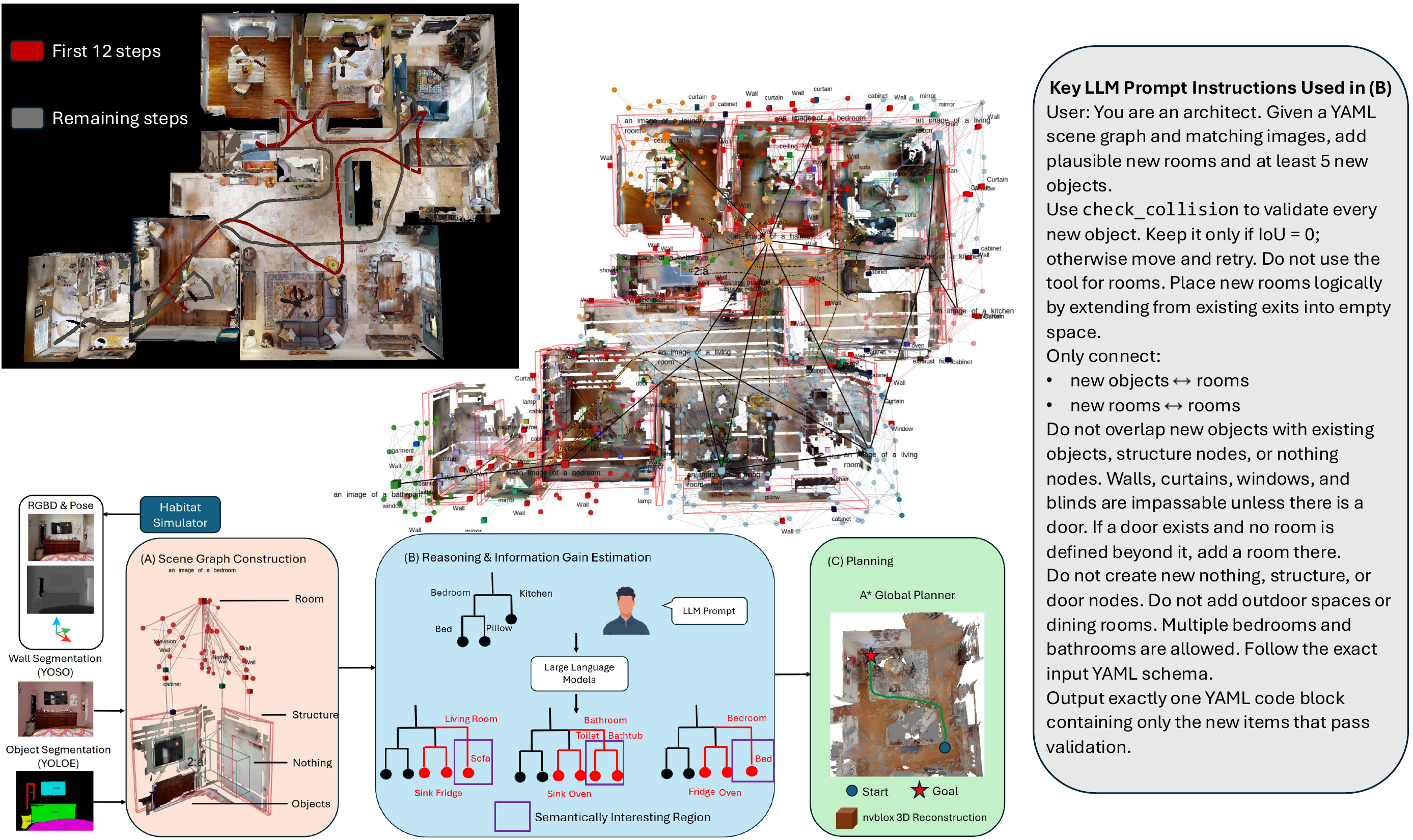}
\caption{
\textbf{Top (left):} A robot exploring an indoor environment using our proposed method. The first 12 exploration steps are highlighted in red, while the remaining steps are shown in gray. The robot quickly covers most semantically salient regions in spite of visual occlusions and narrow corridors/doors, after which it explores the semantic details of the scene.
\textbf{Top (middle):} The final scene graph and mesh constructed after the robot fully explores the environment (visualization using Clio \cite{maggio2024clio}).
\textbf{Bottom:} Our approach consists of three main components. The scene graph construction/mapping module (A) takes RGB-D and pose information from Habitat simulator, object segmentation from YoloE, wall segmentation from YOSO as input, and constructs a multi-layer scene graph with four elements: objects, rooms, structures such as walls, doors or windows and ``nothing'' which represents free space.
The information gain estimation module (B) provides the scene graph and a user prompt as input to an LLM to generate putative scene graphs that are consistent with the current scene graph. The panel shows three such samples, with the most informative region highlighted by the purple rectangle.
The planning module (C) includes a global A* planner that constantly computes a feasible path using an occupancy grid built by nvblox \cite{millane2024nvblox}.
\textbf{Right:} Key LLM prompt instructions used in module (B), including scene expansion, collision checking, graph connectivity, and output formatting constraints.
}
\label{fig:intro_fig}
\end{figure*}

\textbf{Active semantic perception refers to using the semantics of the scene to decide the next viewpoint, so as to achieve tasks like finding a person quickly.}
\cref{fig:intro_fig} gives an overview of our approach.
It has been shown that active perception \cite{bajcsy2018revisiting} requires a robot to (a) maintain a representation of the scene that summarizes past observations and update it with new ones (a map), (b) synthesize new observations of the scene (a generative model), and (c) select controls that maximize information gain (planning) \cite{siming2024active}.

\textbf{The key idea proposed in this paper is to represent semantics of the scene for active perception implicitly in terms of a large language model (LLM) and a scene graph.}
In view of the three desiderata above, there are three key parts to instantiating this idea.
First, we improve upon existing work \cite{hughes2022hydra,maggio2024clio} to develop a compact, multi-layer scene graph that captures rooms, objects, negative space, and structural boundaries such as walls, doors, and windows in indoor environments.
Our graph can represent large, complex environments at various levels of abstraction, e.g., objects inside a room and rooms within a house.
Second, we develop a procedure that uses an LLM to synthesize plausible scene graphs of unobserved regions that are consistent with the partially known scene graph.
This enables an LLM to infer rich semantics and make complex inferences for active perception.
This generative model allows us to rigorously compute an information gain of a potential waypoint directly upon the scene graph that reflects the reduction in the uncertainty of the scene graph: for example, of the two doors in the living room, one could lead to a kitchen and the other to a bedroom.
We develop verification procedures to ensure that these generated plausible scene graphs are also metric-accurate, e.g., the coffee table does not intersect with the couch.
%
Finally, our system also represents fine-grained details of the environment for computing collision-free trajectories that can be executed by the robot.

We experimentally validate these ideas in large simulated indoor environments (with 6--11 rooms) as well as in the real world where a Unitree Go 2 robot explores a typical apartment with six rooms.
On tasks such as detecting objects accurately and exhaustively, or finding different rooms quickly, we show that our approach can discover information about the scene faster and more accurately than baseline methods.
Our system is efficient. Our entire approach, including the autonomy stack, can run on a standard GPU-enabled laptop with API calls to cloud-based LLMs.

%% file: 3-related_works.tex

\section{Related Work}

\textbf{Map Representations:} Classical approaches discretize space and store occupancy probabilities (e.g., an octree) or signed distance functions \cite{oleynikova2017voxblox}. Such maps are largely agnostic to semantics. Recent work augments such maps with per-voxel/per-region annotations of object categories \cite{asgharivaskasi2023semantic}. Such representations are also excessively dense, which makes high-level reasoning difficult.
Neural radiance fields (NeRF) \cite{mildenhall2021nerf} and 3D Gaussian Splatting \cite{kerbl20233d} are photorealistic scene representations. Recent extensions incorporate semantics into these models \cite{semanticnerf, strong2025next}.
These representations can render new views, but they do not represent objects or spatial relations, which makes predictive reasoning about unobserved regions difficult.
In contrast, scene graphs provide a sparse alternative that explicitly represents objects and their spatial relations, and have been used for robot planning \cite{dai2024optimal}. Such methods show that relational scene abstractions can support high-level decision making. However, they typically assume that the scene graph or map is given. In contrast, our work focuses on exploration and the robot must incrementally construct the scene graph from partial observations.

\textbf{Exploration:}
Frontier-based methods navigate to the boundary between free and unknown regions \cite{yamauchi1997frontier,tao2024learning}, while information-driven methods identify new viewpoints/trajectories using expected information gain \cite{bircher2016receding, charrow2015information, he2025active, siming2024active,Jiang2023FisherRF}.
These methods do find unexplored areas but may not necessarily find semantically interesting areas. The reason is as follows. Representations like occupancy grids only encode geometry. Even if some methods represent object categories, they lack the mechanisms to compute semantic information, e.g., to predict that the bedroom could be adjacent to the room with a couch and a coffee table.

\textbf{Goal and Instruction Conditioned Navigation:}
SemExp \cite{chaplot2020object} trains high-level policies to predict long-term goals that are semantically correlated with user-specified tasks, with a local planner for navigation. StructNav \cite{chen2023not} uses LLMs to rank geometric frontiers according to their task relevance, and then passes execution to a local planner. VLFM \cite{yokoyama2024vlfm} constructs a 2D top-down BLIP map from RGB observations, where the BLIP value is computed with respect to a specified target object, and uses this map to score geometric frontiers. While these approaches incorporate semantics, they are designed for locating predefined object categories  and do not reason about semantic uncertainty. Vision-language navigation/action (VLN/VLA) methods condition agents on language instructions or task-specified goals, e.g., natural-language routes \cite{anderson2018vision}, instruction-conditioned navigation policies \cite{zhang2024uni} and end-to-end visuomotor control \cite{kim2024openvla}. These works typically address object retrieval, not efficient exploration. They do not guarantee coverage and efficiency of the path. They do not represent the environment explicitly, and due to the limited memory, they often do not perform exploration tasks reliably \cite{tao2025halo}.

\textbf{Semantic and Graph Completion:}
Semantic scene completion methods predict dense occupancy and object categories of voxels \cite{song2017semantic, cao2022monoscene}. These methods do not infer object-object relationships or objects/rooms in unobserved regions. Graph-based methods can complete partial scene graphs \cite{wan2018representation, garg2021unconditional}. However, they typically operate on static image-level graphs or abstract scene graphs. They do not complete metrically grounded, hierarchical scene graphs with new rooms, objects, relations, and spatial locations. In contrast, our method hypothesizes new semantic entities and their locations from partial observations of the scene.

%% file: 4-method.tex

\section{Methods}

\subsection{Problem Description and Overall Approach}
We address the problem of robotic autonomous exploration of an unknown indoor environment to detect and localize all objects (tables, chairs, couch, beds, TV, etc.), high-level entities such as rooms (living room, bedroom, kitchen, bathroom etc.), and structural information (walls, doorways, windows, etc.) while minimizing the traversed path length. We are interested in systems that can utilize semantic information in the environment to perform such exploration. We will therefore evaluate different approaches in terms of (i) how accurately and quickly they detect these entities above, and (ii) how well they can predict unseen structure in the environment before actually observing it.

Our approach, illustrated in \cref{fig:intro_fig}, has three key parts.
(i) The mapping module takes in RGB-D, segmentation images, and localization poses, and outputs an ensemble of scene graphs.
(ii) The reasoning and information gain estimation modules use these scene graphs to sample semantically consistent potential completions and identify regions with the highest information gain.
(iii) The planning module generates a collision-free path for the robot to navigate to these regions.

\subsection{Mapping}
Our scene graph uses two types of nodes from Clio. It also adds two new node types that improve reliability of the scene graph and enrich the semantics.

\textbf{Object nodes:}
To ensure both speed and accuracy of detection, we choose a unified and single stage architecture, YOLOE \cite{wang2025yoloe}, instead of FastSAM \cite{zhao2023fast} and CLIP \cite{radford2021learning} based approach used in Clio. In our experience, this leads to better recall. We construct a 3D voxel map (based on the depth of the pixels) for every segmentation mask. We also maintain a set of active tracks for each object. This allows us to greedily associate new observations to the tracks by computing the intersection over union (IoU) of the voxels in the current detection and voxels in the track. An association is successful only when the categories of the objects match. A new track is spawned if no track can associate with a detected object. Finally, if a track has not been associated for $\tau$ seconds, it is terminated. We then use the marching cubes \cite{lorensen1998marching} algorithm to fit a mesh for every track. This ensures tight bounding boxes for all objects.

\textbf{Room nodes:}
The following steps are similar to those in Clio. We construct a truncated signed distance function (SDF) for the background and compute a Euclidean SDF from it using the brushfire algorithm in Voxblox \cite{oleynikova2017voxblox}. We extract the sub-graph of places using a generalized Voronoi diagram built on-the-fly while computing the ESDF. The feature vector of a ``place'' node is the average of CLIP embeddings of images in which the node is visible. We add occlusion checking to the original Clio implementation to make these features more accurate. Next, we apply agglomerative clustering to group multiple ``place'' nodes into higher-level ``room'' nodes. The feature vector of a ``room'' is the average embedding of its constituent ``place'' nodes, and its position is equal to their average positions. Finally, we assign a ``room'' label by matching the aggregated feature to the closest predefined ``room'' category.

\textbf{``Nothing'' nodes:}
A scene graph is inherently sparse, and constructing a realistic and informative representation of the environment for our tasks requires more than simply modeling objects and rooms. RGB-D images and segmentation results not only reveal where objects are present, but also where no objects exist in the scene. The absence of objects is informative. We encode this negative space by introducing a ``nothing'' object, which explicitly represents regions confirmed to be free of objects. We threshold the TSDF to get the 3D occupancy grid and identify the largest free-space cuboid within it. This is the classic problem of finding the ``largest rectangle in a histogram'', extended to three dimensions.

\textbf{Structure nodes:}
In the scene graph, ``room'' nodes only store positional information and room labels. To capture more detailed spatial properties, we introduce structure nodes that encode the size and shape of each room. Our technique follows the VS-Graph framework \cite{tourani2025vs} with YOSO \cite{hu2023yoso} as the detection front-end. The structure node incorporates elements such as walls, doors, windows, blinds, and curtains. After segmentation, we first perform RANSAC \cite{fischler1981random} plane fitting on the point cloud data associated with each segmentation mask. We then fit a bounding box to all the inliers. Walls are quite difficult to detect accurately. So we require that each wall be observed multiple times before being added to the graph. We further apply Euclidean clustering to the RANSAC output to prevent the generation of excessively long walls. This was important to ensure that the scene graph does not merge spatially separated but co-planar wall segments into a single excessively long wall.

\subsection{Reasoning and Information Gain Estimation}
Let $x_k \in \text{SE}(2)$ denote the location of the robot at time $k$ and $Y_k$ denote the observation at time $k$. The realization of this random variable will be denoted by $y_k$.
The goal of active perception \cite{siming2024active} is to select a next viewpoint $x_{k+1}$ that maximizes the mutual information between the next observation $Y_{k+1}$ and past observations $(y_1,\dots,y_k)$
\beq{
    \aed{
    \text{I} (Y_{k+1}, y_{1:k}\mid x)
    &= \text{I} (Y_{k+1}, G_k \mid x)\\
    &= \text{H}(Y_{k+1} \mid x) - \text{H}(Y_{k+1} \mid x, G_k).
    }
    \label{eq:I}
}
The first equality follows because the current scene graph, denoted by $G_k$, is a representation of the past observations. The second equality follows by the definition of mutual information with $\text{H}(\cdot)$ being Shannon entropy.
The second term $\text{H}(Y_{k+1} \mid x, G_k)$ is the entropy of the conditional distribution $p(Y_{k+1} \mid x, G_k)$ averaged over the scenes $G_k$. It equals
\beq{
    -\int \dd{p(G_k)} p(Y_{k+1} \mid x, G_k) \log p(Y_{k+1} \mid x, G_k).
    \label{eq:h_y_x_g}
}
The first term is the entropy of the marginal distribution $p(Y_{k+1} \mid x) = \int \dd{p(G_k)} p(Y_{k+1} \mid x, G_k)$.
We therefore need to synthesize observations $Y_{k+1}$ from putative viewpoints $x$ to calculate the information gain.
We split this into two problems: (i) sample a completed scene graph $G$ that is consistent with the current graph $G_k$, and (ii) render an observation $Y_{k+1}$ from this sampled scene graph $G$.
Observe that the information gain does not depend upon the photometric details in $Y_k$ if the scene graph does not store these details.

\textbf{Sampling a completed scene graph}
that is consistent with $G_k$ involves sampling from
\(
    p(G \mid G_k).
\)
This requires knowing both typical layouts of scenes and having the ability to generate new ones.
LLMs are an excellent fit here due to their rich language priors.
We provide the scene graph to the LLM in two different formats:
(i) a YAML file that contains all the numerical information for different layers of the scene graph, and
(ii) a visual representation of the scene as images of cross-sections at various heights (e.g., heights [0.0, 0.3] or [0.3, 0.5]).
We instruct the LLM to expand the current scene graph by proposing plausible new rooms and objects. To ensure physical consistency, we provide the LLM with a collision checking tool over existing nodes in the graph and require it to validate each proposed node before adding it to the graph.
This procedure is used to sample an ensemble of graphs $\{G^{(i)}\}_{i=1}^m$. We can now approximate:
\beq{
    p(G \mid G_k) \equiv \f{1}{m} \sum_{i=1}^m \delta_{G^{(i)}}(G).
    \label{eq:p_g_gk}
}
where $\delta$ is the Dirac delta distribution.

\begin{figure}[ht]
    \centering
	\includegraphics[width=0.9\linewidth]{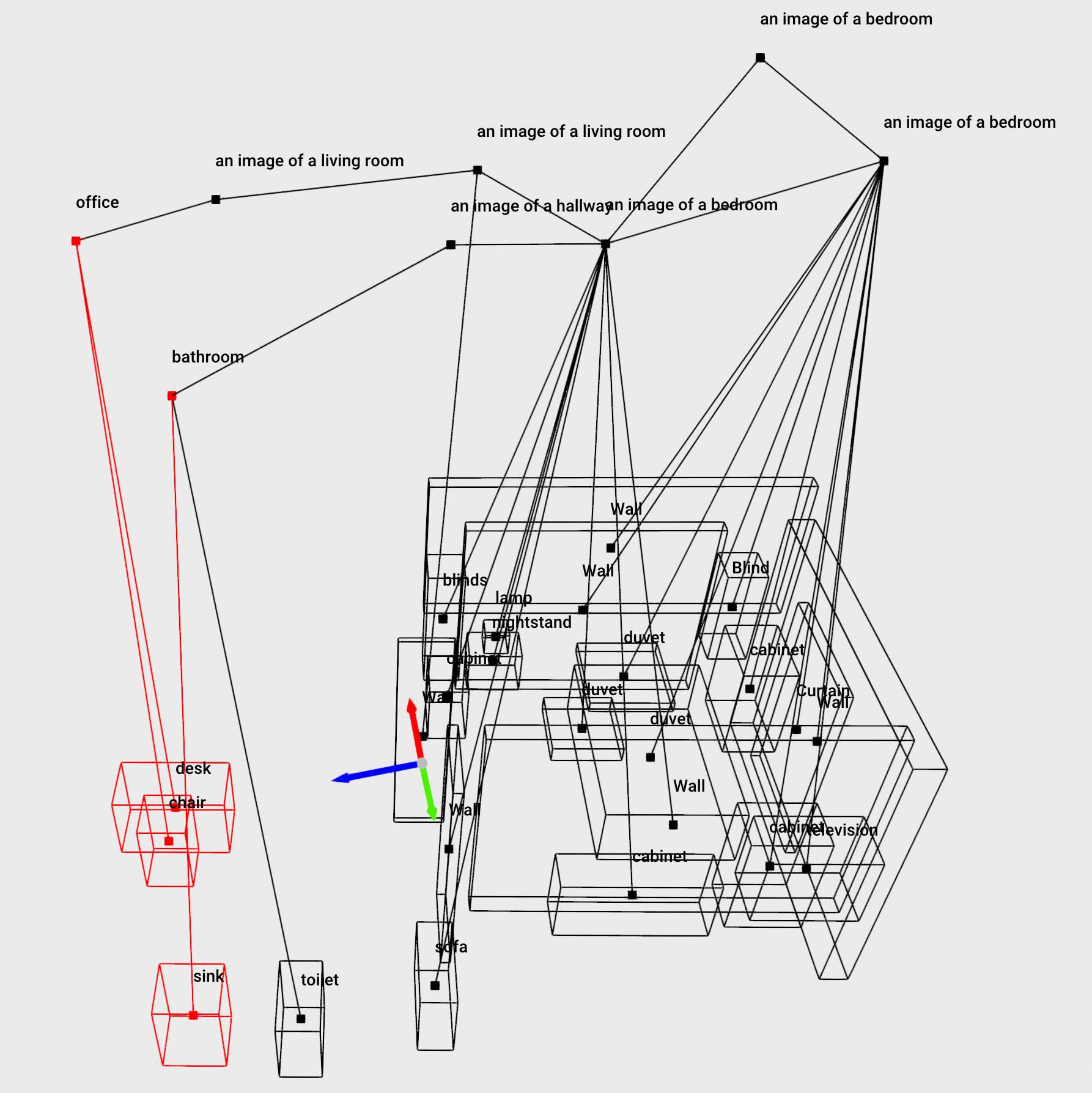}
	\caption{The LLM produces a semantically plausible completion of a given scene graph. It places an office and a bathroom adjacent to the bedroom while maintaining reasonable separation between them. The geometry of the completion is also broadly consistent with real-world layouts. For example, the sink is placed next to the toilet. The red portion of the scene graph indicates what the camera is expected to observe from a particular viewpoint.
	}
	\label{fig:future_obs}
\end{figure}

\textbf{Rendering future observations}
Unlike a NeRF, a scene graph does not automatically render an observation from novel views. We define the observation of scene graph $G$ from a viewpoint $x$, denoted by $Y(G, x)$, to be (i) the set of nodes in $G$ (across all layers) that are visible from the viewpoint (i.e., at least partially within the field of view of the camera) and not occluded by any structure nodes (e.g., walls, doors, windows), and (ii) all the ``room'' nodes associated with such nodes.
Occlusions of object nodes by structure nodes are checked by ray tracing from the camera viewpoint to the object centroid.
\cref{fig:future_obs} shows an example observation rendered from a scene graph completed by an LLM.
Under this model, the observation $Y(G, x)$ is a deterministic function of the graph $G$ and viewpoint $x$. Observe that
\[
p(Y_{k+1} \mid x) =  \frac{1}{m} \sum_{i=1}^m \delta \rbr{Y_{k+1} - Y(G^{(i)}, x)}.
\]
We next model the uncertainty of this observation to compute the entropy $\text{H}(Y_{k+1} \mid x, G)$.%
\footnote{This is easy to do in a NeRF where we can compute both the image and its variance using the rendering equation \cite{siming2024active}.}
We perturb the nodes of $G$ to modify their locations. Let these perturbations be denoted by $\e$. Perturbations are chosen to be consistent with the sizes of nodes, e.g., objects are perturbed within the room (their parent node). This allows us to compute
\beq{
p(Y_{k+1} \mid x, G) =  \frac{1}{N} \sum_{i=1}^N \delta \rbr{Y_{k+1} - Y(G, x, \e^{(i)})},
\label{eq:p_y_x_g}
}
for observations $Y(G, x, \e^{(i)})$ from perturbed scene graphs. This distribution can now be used to compute $\text{H}(Y_{k+1} \mid x, G)$.
To capture scene uncertainty at multiple semantic levels, we compute the information gains separately for ``object'' nodes and ``room'' nodes, and combine them into a single score for each viewpoint $x$. To prioritize nearby regions, we add a penalty for the travel distance. The final score is
\beq{
    \text{I}(x) = \text{I}_{\text{object}}(x) +\l_r \text{I}_{\text{room}}(x) - \l_d d.
    \label{eq:ig}
}
where $d$ is the Euclidean distance between $x$ and the robot's current position.%
\footnote{When computing information gain for rooms, we include only the dominant room nodes (those containing most objects in the observation) rather than all room nodes. This avoids selecting viewpoints at the boundary between two rooms.}

In practice, some viewpoints that are highly informative under the completed scene graph may not be physically reachable in the occupancy map, e.g., narrow corridors could be blocked due to mapping errors. To account for this, we first select the top-$k$ viewpoints according to \cref{eq:ig}. For each candidate, we plan a path in the occupancy grid and redefine the putative viewpoint as the farthest reachable point along the planned path that faces the original target region. We recompute the information gain at this new viewpoint and use it to re-rank the original candidates. The target region itself is not changed during this procedure.%
\footnote{This was a useful heuristic: for highly informative target views, the robot might discover a more accurate map and thereby find a viable path to the target as it traverses to this high-scoring intermediate view.}

\subsection{Planning}
We construct an occupancy grid from RGB-D observations using nvblox \cite{millane2024nvblox}. If a goal lies outside the occupancy grid or does not fall within known free space, we project it onto the nearest free voxel and use A* to plan a path to that location. As new observations are incorporated into the occupancy grid, we continuously re-plan to reach the target. Velocity commands along trajectories are executed by the robot using the Unitree navigation SDK.

%% file: 5-experiments.tex

\section{Experimental Validation}

We next experimentally validate our approach in large simulated indoor environments as well as the real world where a Unitree Go 2 robot explores an apartment with six rooms. Our goal is to ascertain whether the system detects objects accurately and exhaustively, and finds different rooms in the scene quickly. We also discuss a number of experimentally observed behaviors that shed light on active perception methods.

\textbf{Baselines:}
We compare against:
(i) frontier-based exploration \cite{yamauchi1997frontier} that identifies unexplored frontiers on an occupancy grid without semantic awareness, and
(ii) SSMI \cite{asgharivaskasi2023semantic} which incorporates semantic information in the occupancy grid and evaluates geometric frontiers to rank paths based on the estimated reduction of semantic uncertainty.
For all methods, we perform a 360° yaw rotation at the end of each exploration step to maximize visual information.
All methods are run for a maximum path length of $L_{\text{max}}$. Frontier exploration can terminate early if it does not detect open frontiers.
We also record the time spent in collecting observations from the scene (this includes time spent in traveling and in-place yaw rotations). For comparing these approaches, we exclude the time spent in computing the planned trajectory or querying the LLM.

For our approach, we use Gemini-2.5-Pro \cite{comanici2025gemini}.%
\footnote{This is an expensive model (each query costs about \$0.16 for us). One step of our approach (which selects a target location for the robot) costs \$1.28. A full experiment has 25--30 steps, so the cost is high but not prohibitive. In our experience, smaller, less expensive LLMs do not yet provide consistent enough performance to support a fully autonomous robot.}
We build two scene graphs on bootstrapped subsets of sensory observations to represent $p(G_k)$ in \cref{eq:h_y_x_g}. We sample $m=4$ completions in \cref{eq:p_g_gk} using the LLM for each constructed scene graph. This yields 8 samples for the completed graph $G$. For calculating the information gain, we perturb each graph $N=4$ times in \cref{eq:p_y_x_g} and randomly sample 300 candidate viewpoints distributed uniformly throughout the scene. We set $\lambda_r = 1,\ \lambda_d=0.2$ in \cref{eq:ig}.

\subsection{Simulation experiments}

We conduct simulation experiments on four representative scenes from the HM3D \cite{ramakrishnan2021hm3d} dataset ranging from small to large (scene 1 id: 853; scene 2 id: 69; scene 3 id: 871; scene 4 id: 573). Navigation is performed in the Habitat simulator \cite{savva2019habitat} on a laptop with an 8-core CPU and an Nvidia 4060 GPU. To construct ground-truth scene graphs for evaluation, we manually navigate the camera viewpoint through each environment, ensuring that all accessible regions are visited and all objects are detected.

\begin{figure}
    \vspace{4pt}
	\centering
        \includegraphics[width=0.64\linewidth]{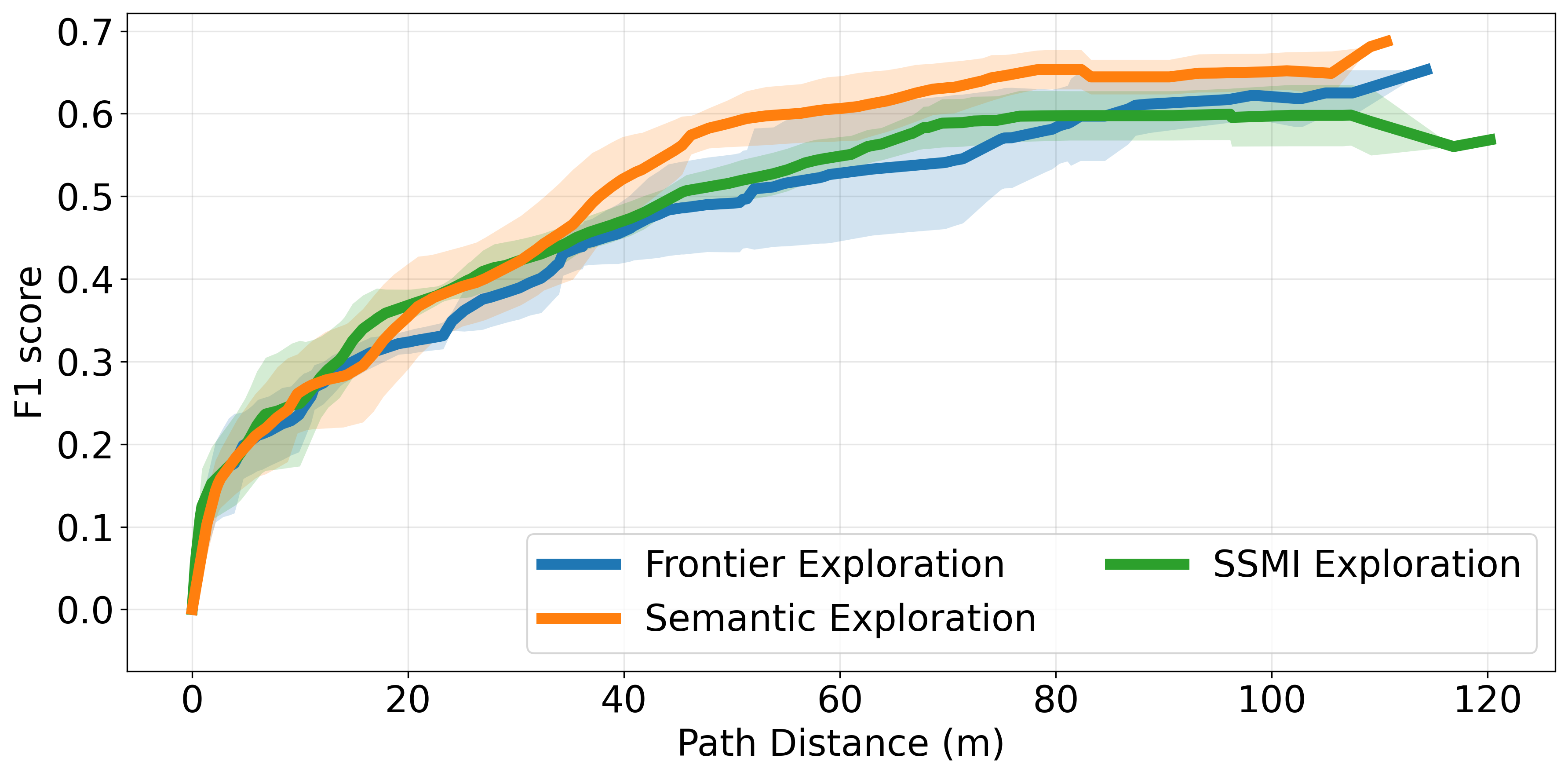}
        \vspace*{1ex}
        \includegraphics[width=0.64\linewidth]{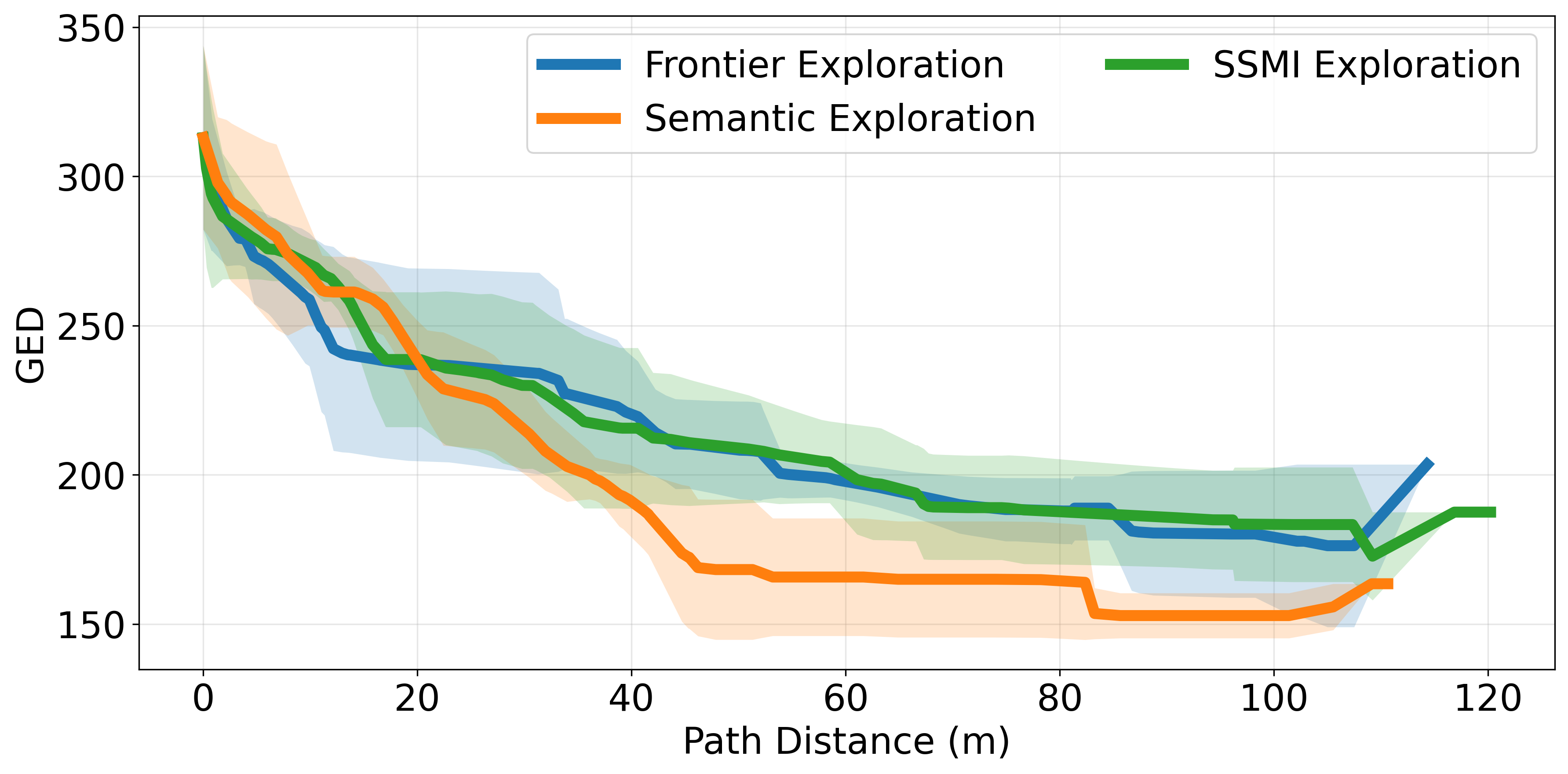}
	\caption{\textbf{Our approach has a higher average F1 score and lower GED compared to baselines for different path lengths.} The mean travel distance per step is 2.76 m for Frontier, 3.74 m for our approach and 4.04 m for SSMI.
	}
	\label{fig:semantic_ground}
\end{figure}

\textbf{Semantic grounding capability:}
This experiment evaluates how quickly and effectively an agent can build a semantic understanding of its environment. We assess the quality of the constructed scene graphs using two complementary metrics: F1 score and graph edit distance (GED). The F1 score captures the accuracy of object detection by measuring the proportion of correctly identified objects across 61 common indoor object categories, including furniture, household appliances, fixtures, decor items, and everyday personal items.%
\footnote{We consider two objects a match if they have the same category and the distance between their centroids is within 0.5 m. We restrict this calculation to object nodes. Structure and nothing nodes which serve to constrain the space of scene graphs are excluded.}
In contrast, GED evaluates topological correctness, including rooms, objects, and their relationships. GED between two graphs is the minimum sequence of node and edge edit operations required to transform one graph into a graph isomorphic to the other.%
\footnote{In our implementation, insertion and deletion of nodes and edges are assigned equal cost (1), ensuring balanced treatment of all graph components. Substitution is permitted only under specific conditions: room nodes must share the same name, while object nodes must share both the same name and the same parent, with spatial proximity within defined thresholds (0.5 m for objects and 4.0 m for rooms).}
\cref{fig:semantic_ground} shows that, on average across the four scenes, our approach has a better F1 score and GED for different path lengths. The variance is smaller at the end because exploration of different scenes terminates at different distances. We also report F1 and GED at 10\%, 50\%, and 100\% of a fixed exploration time budget in \cref{fig:quantitative} (left). This shows that our approach can quickly build a topologically consistent semantic understanding of the scene.

\begin{figure*}
\centering

\begin{minipage}[c]{0.56\linewidth}
\fbox{
\huge
\begin{adjustbox}{width=\linewidth}
\renewcommand{\arraystretch}{1.6}
\begin{tabular}{l rr rr rr p{2ex} rr rr rr}
& \multicolumn{2}{c}{10\%} & \multicolumn{2}{c}{50\%} & \multicolumn{2}{c}{100\%}
&& \multicolumn{2}{c}{10\%} & \multicolumn{2}{c}{50\%} & \multicolumn{2}{c}{100\%}\\
\cmidrule(lr){2-3}\cmidrule(lr){4-5}\cmidrule(lr){6-7}
\cmidrule(lr){9-10}\cmidrule(lr){11-12}\cmidrule(lr){13-14}
& F1$\uparrow$ & GED$\downarrow$ & F1$\uparrow$ & GED$\downarrow$ & F1$\uparrow$ & GED$\downarrow$
&& F1$\uparrow$ & GED$\downarrow$ & F1$\uparrow$ & GED$\downarrow$ & F1$\uparrow$ & GED$\downarrow$\\
\toprule
\addlinespace\addlinespace
& \multicolumn{6}{c}{\textbf{Scene: 573}} & \multicolumn{6}{c}{\textbf{Scene: 853}}\\
\addlinespace\addlinespace
Frontier
& \textbf{19.9} & \textbf{231.4} & 46.1 & 193.0 & 63.0 & 185.0
&& 18.6 & 279.0 & 46.5 & 243.4 & 58.8 & 177.0\\
SSMI
& 12.2 & 261.0 & 52.1 & 187.5 & 56.8 & 187.5
&& 28.5 & 280.3 & 49.5 & 218.0 & 61.1 & 189.0\\
Semantic (Ours)
& 19.2 & 252.6 & \textbf{59.0} & \textbf{170.5} & \textbf{68.7} & \textbf{163.5}
&& \textbf{34.4} & \textbf{249.5} & \textbf{60.8} & \textbf{147.0} & \textbf{66.7} & \textbf{147.0}\\

\addlinespace\addlinespace
& \multicolumn{6}{c}{\textbf{Scene: 871}} & \multicolumn{6}{c}{\textbf{Scene: 69}}\\
\addlinespace\addlinespace
Frontier
& 29.2 & 292.1 & 55.7 & 233.0 & 65.3 & 203.5
&& 23.8 & \textbf{253.0} & 40.6 & 213.0 & 59.7 & 149.0 \\
SSMI
& \textbf{33.4} & \textbf{277.3} & 59.1 & 212.4 & 61.7 & 204.5
&& \textbf{25.9} & 269.2 & 50.0 & 194.0 & \textbf{63.1} & 158.0 \\
Semantic (Ours)
& 30.5 & 279.0 & \textbf{62.4} & \textbf{204.5} & \textbf{68.0} & \textbf{194.5}
&& 25.8 & 266.3 & \textbf{60.1} & \textbf{148.0} & 62.2 & \textbf{148.0} \\
\bottomrule
\end{tabular}
\end{adjustbox}
}
\end{minipage}%
\hspace*{2em}
\begin{minipage}[c]{0.33\linewidth}
\vspace{4pt}
\fbox{
\begin{adjustbox}{width=\linewidth}
\renewcommand{\arraystretch}{1}
\centering
\begin{tabular}{l rrrr}

& $T_{\text{pred}}(s) \downarrow$ & $D_{\text{pred}}(m)\downarrow$ & $T_{\text{find}}(s) \downarrow$ & $D_{\text{find}}(m) \downarrow$\\
\toprule
\addlinespace
\multicolumn{5}{l}{\textbf{Scene: 853, 6 rooms, 207.63 m$^2$. Target: second bedroom}}\\
Frontier    & - & - & 200  & 25\\
SSMI & - & - & 345 & 61 \\
Semantic (Ours) & \textbf{65} & \textbf{14} & \textbf{173} & \textbf{32} \\

\addlinespace
\multicolumn{5}{l}{\textbf{Scene: 69, 7 rooms, 206.42 m$^2$. Target: bedroom}}\\
Frontier & - & - & 56  & \textbf{5}\\
SSMI  & - & -& 42 & 7 \\
Semantic (Ours) &  \textbf{0} & \textbf{0} & \textbf{33} & 6 \\

\addlinespace
\multicolumn{5}{l}{\textbf{Scene: 69, 7 rooms, 206.42 m$^2$. Target: kitchen}}\\
Frontier  & - & - & 558  & 87\\
SSMI & - & - & 409 & 61 \\
Semantic (Ours) &  \textbf{233} & \textbf{39} & \textbf{291} & \textbf{45} \\

\addlinespace
\multicolumn{5}{l}{\textbf{Scene: 871, 9 rooms, 235.03 m$^2$. Target: kitchen}}\\
Frontier  & - & - & 258  & 31\\
SSMI & - & - & \textbf{96} & \textbf{14} \\
Semantic (Ours) &  \textbf{89} & \textbf{14} & 150 & 21 \\

\addlinespace
\multicolumn{5}{l}{\textbf{Scene: 573, 11 rooms, 537.81 m$^2$. Target: bathroom}}\\
Frontier    & - & - & 34  & \textbf{5}\\
SSMI &  - & - & 47 & 7 \\
Semantic (Ours) &  \textbf{7} & \textbf{2} & \textbf{27} & 6 \\
\bottomrule
\end{tabular}
\end{adjustbox}
}
\end{minipage}
\caption{
\textbf{Left:} Quantitative results of active exploration in four simulation environments. F1 score and graph edit distance (GED) are evaluated at 10\%, 50\%, and 100\% of the total navigation time.
\textbf{Right:} Quantitative results for the ability to predict the existence of new rooms in simulated apartment scenes from Habitat before actually finding them during exploration. The quantities $T_\text{pred}$ and $D_\text{pred}$ denote time and distance respectively during exploration when such a prediction was made while $T_\text{find}$ and $D_\text{find}$ denote time and distance when the room was actually found.
\label{fig:quantitative}
}
\end{figure*}

\textbf{Ability to reliably predict new, as yet unobserved, rooms:}
Beyond identifying semantically uncertain regions, a key strength of our approach lies in its ability to complete partially observed scenes. Since the completed scene graph is conditioned on the observed scene graph, different regions of the scene carry different levels of uncertainty. Certain parts of the scene can be inferred with high confidence in the LLM ensemble, while others remain ambiguous. To evaluate this capability, we focus on discovering rooms that the robot has not directly observed. For this experiment, we will take the completed graph as the current scene graph augmented with confident predictions. A prediction is considered confident if it appears in at least two of the eight ensemble samples.

\begin{figure*}[!t]
	\centering
        \includegraphics[width=0.75\linewidth]{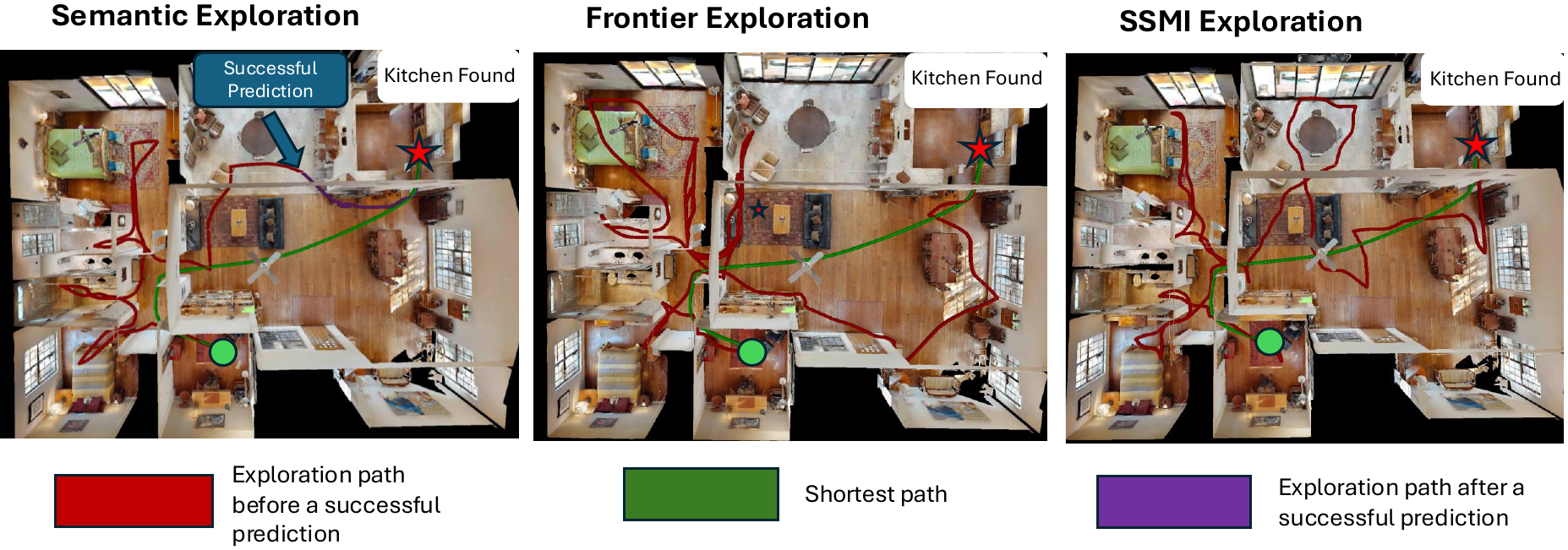}
    \caption{\textbf{Paths taken to locate a kitchen:} Semantic Exploration (ours) identifies a hole in the wall and predicts the location of the kitchen across it before reaching the kitchen. Consequently, it has a shorter trajectory to the target than the baseline methods.
    }
	\label{fig:room_find}
\end{figure*}

\cref{fig:quantitative} (right) reports these results. Here, $T_{\text{pred}}$ and $D_{\text{pred}}$ denote the time and distance required for a successful prediction, while $T_{\text{find}}$ and $D_{\text{find}}$ denote the time and distance to actually discover the target room (during the course of navigation). This analysis is performed post-hoc on recorded data after termination. Success requires satisfying two conditions: (i) the room appears in at least two of the eight ensemble outputs, and (ii) the Euclidean distance between the predicted and actual room locations must be within 2.5 meters. We also visualize the trajectories of all methods in \cref{fig:room_find}. Our method is able to predict potential target rooms before they are physically discovered by any method. Our method also obtains a better final discovery time and distance in many, but not all, cases. This behavior is interesting because we do not tell the LLM which target room to find in the prompt. We have checked that for about 30\% of the rooms across these four scenes (9 out of 28), our system can successfully predict a new room before the robot finds it.

\begin{figure*}
    \centering
	\includegraphics[width=0.75\linewidth]{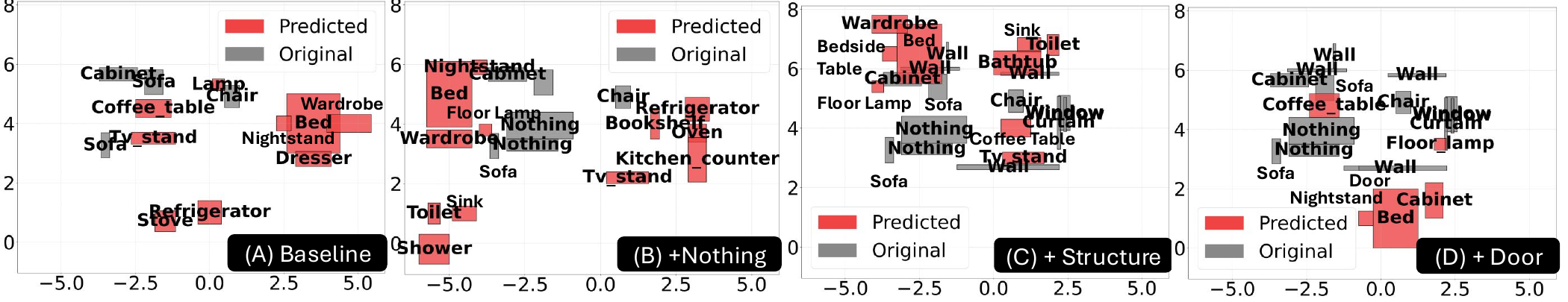}
	\caption{
	\textbf{Marked improvements in the plausible scene graph as node types are enabled progressively.}
	}
	\label{fig:ablation}
\end{figure*}

\textbf{Ablation study:}
We next evaluate the contribution of each node type on a single scene, while progressively enabling each type: (A) Baseline nodes; (B) Baseline + Nothing (negative-space) nodes; (C) Baseline + Nothing + Structure nodes; and (D) Baseline + Nothing + Structure + Door nodes. \cref{fig:ablation} overlays predicted boxes (red) on existing boxes (gray). Relative to (A), adding Nothing nodes in (B) reduces incorrect object placement (e.g., a TV stand) by explicitly encoding free space. Adding Structure nodes in (C) constrains completions to respect room boundaries and reduces proposals being outside the enclosed space or the window. Adding Door nodes in (D) guides new-room proposals behind doors where a new room is most likely to be found.

\subsection{Discussion}

\textbf{Exploration behavior:}
Our method has a large mean travel distance per exploration step. It effectively ``sees farther'' than baseline methods. Semantically interesting locations are typically far away from the robot. We see in our quantitative and qualitative results that our method reliably finds such locations---much like the firefighters in the introduction would. It infers the structural layout inside known rooms easily and therefore ignores such rooms for exploration. Different methods perform roughly the same at the beginning, this is because of our trick of a 360° rotation after each step. Our method works better (higher F1, lower GED) at later stages as it uncovers more context to identify more strategic viewpoints than those in free unexplored space or ones with poor visibility. We also noticed that our method sometimes revisits mapped areas to correct previous mistakes.

\begin{figure*}[!htpb]
    \vspace{4pt}
	\centering
	\includegraphics[width=0.75\linewidth, trim={0cm 2.0cm 0cm 1.0cm}]{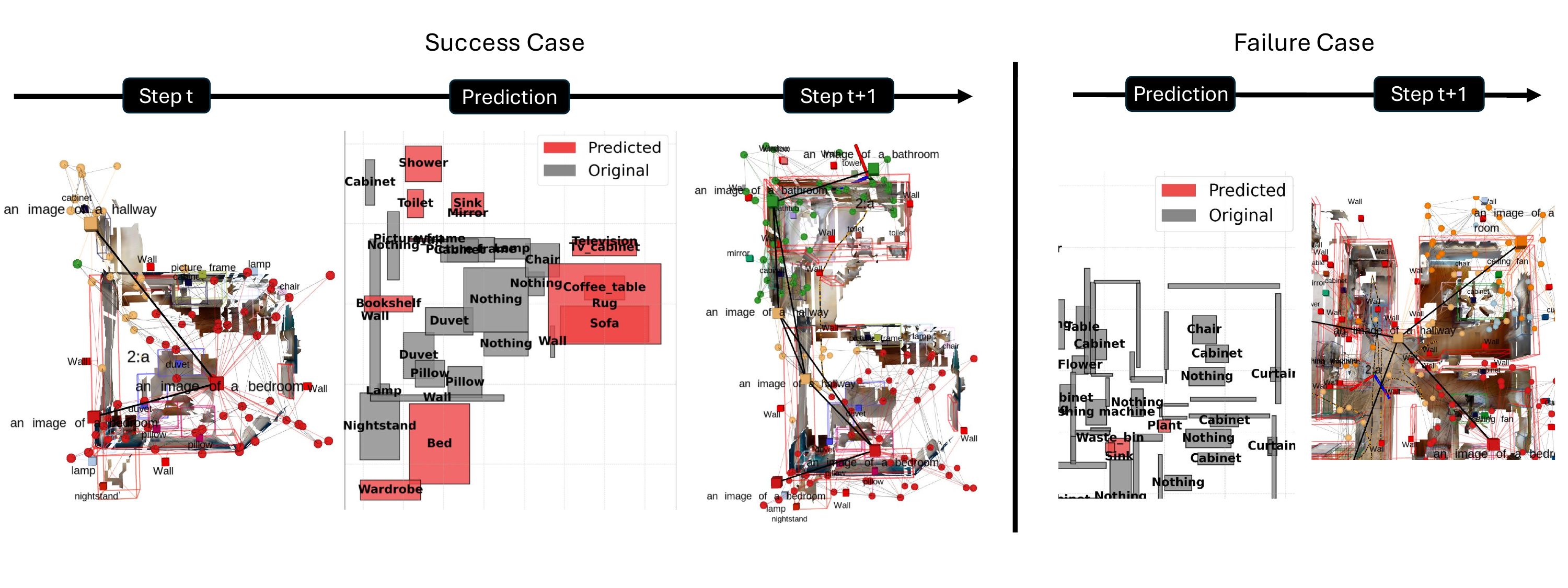}
	\caption{\textbf{Left:} At step $t$, our system successfully predicts a bathroom above the bedroom for the robot to navigate to at time $t+1$.
	We show 2D projections of the scene graph (gray) and predictions (red).
	\textbf{Right:} The system incorrectly predicts a bathroom in the hallway, and misleads the robot to navigate towards an empty region.
	}
	\label{fig:complete_show}

    \includegraphics[width=0.75\linewidth]{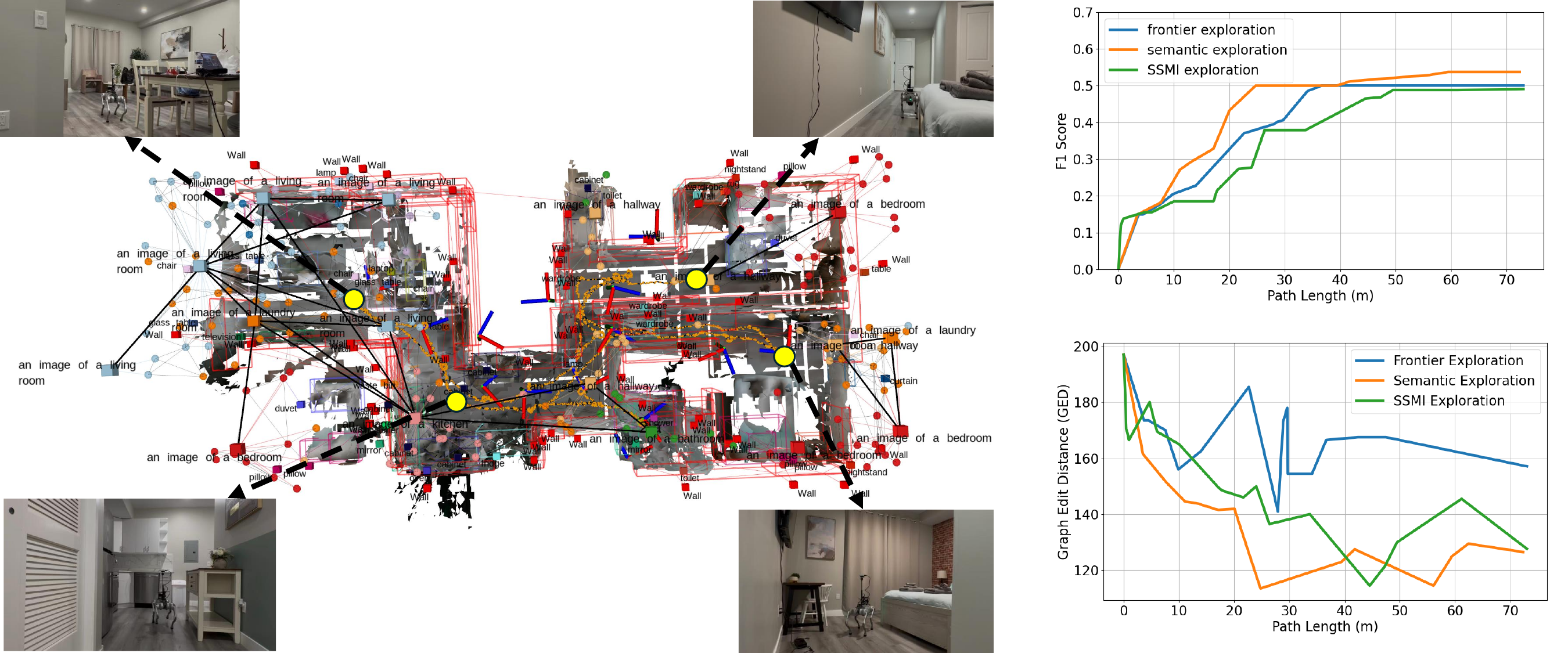}
    \caption{\textbf{Active semantic perception on a Unitree Go 2 robot operating autonomously in a large apartment.} \textbf{Left: } Reconstructed scene graph and some viewpoints selected by our algorithm (visualized as coordinate frames) on the scene graph.
    Yellow markers are third-person snapshots.
    \textbf{Right:} F1 score and GED as a function of path length in the real-world experiment.
    The robot operated autonomously for 55 mins and explored about 1,200 sq.\@ ft. of a congested apartment with six rooms during this experiment. Of these, about 25 mins were spent in navigation, and the rest were spent in reasoning.
    }
    \label{fig:real_world_vis}
\end{figure*}

We observed that SSMI sometimes favors farther geometric frontiers over closer ones. While this does improve exploration, it often results in circuitous trajectories that come back to the same location later. We believe the reason for this is that SSMI uses object categories---these are not semantics of the scene as we argued in the introduction. Noise in the object detector can produce many uncertain voxels. SSMI prioritizes viewpoints on distant frontiers where this uncertainty is typically high.
Our method reasons with both scene and object level semantics. This is perhaps the key reason it obtains shorter and more purposeful paths.

\textbf{Prediction behavior:}
We find that hallways and doors serve as strong cues for the LLM to hypothesize additional rooms. The left part of \cref{fig:complete_show} shows this. At exploration step $t$, the robot has mapped a bedroom and a hallway; conditioned on these cues, the LLM predicts a bathroom near the hallway, a living room to the right, and a bedroom at the bottom. The next viewpoint selected by our information-gain objective lies at the top. At step $t+1$, the robot uncovers a bathroom at the predicted location, confirming that hypothesis (other proposals require further evidence). Hallways can mislead the LLM. The right part of \cref{fig:complete_show} is an example of this. Although the hallway had already been mapped and was simply empty space, our free space representation is not perfectly exhaustive, leaving small gaps between neighboring nothing objects. These gaps allow certain proposals to pass collision checking. The LLM should recognize that such confined regions are unlikely to contain new rooms. It infers a semantically reasonable but physically infeasible bathroom. This causes the robot to revisit an uninformative region.

\subsection{Real-world experiments}

We deploy our method on a Unitree Go 2 robot equipped with an Intel RealSense D435i camera.%
\footnote{The camera is positioned at a height of 0.9 m and tilted downward by 10 degrees to improve occupancy-grid estimation.} Experiments are conducted in a large apartment consisting of six rooms. Robot pose is estimated using RKO-LIO \cite{malladi2025robust} which fuses measurements from a Hesai LiDAR on the robot and the onboard IMU. RGB-D images are streamed at 15\,Hz and pose estimates at 20\,Hz over 5\,GHz WiFi to a laptop for processing. The laptop runs the perception, mapping, planning, and our method. It sends velocity commands to the robot.

The robot runs ROS~2 Foxy with the Unitree ROS~2 SDK while the laptop runs our code in ROS~1 through a ROS~2--ROS~1 bridge. We use Zenoh for wireless communication instead of CycloneDDS for higher throughput and low-latency transmission. All experiments are executed autonomously, with a human operator present only for monitoring safety.%
\footnote{To ensure consistent initial conditions across methods, we place a fixed sheet of paper at the start location and use it to align both the robot's center position and heading before each trial.}
For safe operation, we perform collision checking in two stages: first we check an oriented rectangular footprint against the occupancy grid along each planned path, and then check a circular footprint at the target location where the robot performs a 360° rotation.

\textbf{Results}
\cref{fig:real_world_vis} (left) shows the reconstructed scene graph together with the sequence of viewpoints selected by our algorithm. Despite sensing noise and localization errors, our pipeline can construct a complete and semantically consistent representation of the environment. We also compare F1 and GED against baseline methods, as shown in \cref{fig:real_world_vis} (right). Our method achieves higher F1 and lower GED over most of the trajectory, indicating that the performance gains observed in simulation largely carry over to the real-world setting.

\textbf{Discussion}
The RealSense camera has a relatively narrow field of view (FOV). Our method is resilient to this because it selects target viewpoints that provide more complete observations of the scene. Baseline methods struggle due to this limited FOV.
The camera is mounted with a 10° downward tilt. This reduces the effective range but gives multiple observations for voxels in the vicinity of the robot. This is useful to correct voxels incorrectly marked as occupied (due to latencies and vibrations as the robot tilts during its gait) which can block narrow doorways/corridors.
Exploration becomes more sensitive to the quality of the occupancy grid in parts of the apartment with narrow corridors where the robot cannot perform a full 360° turn. Path planning may fail for certain frontier candidates, which causes baseline methods to exclude potentially informative regions. This severely affects baselines that do not use nvblox \cite{millane2024nvblox}. To prevent such occupancy grid errors from dominating our comparisons, we manually assisted the affected baselines about 2--3 times in each trial.
In contrast, our method runs fully autonomously without any manual intervention.

%% file: 6-future_work.tex

\section{Limitations and Future work}
While this paper demonstrates the impressive potential of semantic exploration, several limitations remain.
The LLM we use is not trained specifically for scene expansion. As a result, its proposal distribution can diverge from the true scene distribution, particularly in larger and more complex environments.
Due to this, we need to make multiple queries to the LLM and prune bad predictions. In our experiments, each iteration, which ends with selecting a distant location to go to, requires about 70 seconds and \$1.28.
Nearly all of the latency is due to queries to the cloud-based LLM.
It is therefore of interest to build LLMs that are tailored to physical scenes and can run locally onboard the robot.
Another promising direction is to extend our framework to handle a broader range of object categories, such as structural exits, and people, thereby improving its applicability to more real-world semantic navigation tasks.

%% file: 7-appendix.tex

%% file: pratik.bib
@article{bajcsy2018revisiting,
  title = {Revisiting Active Perception},
  author = {Bajcsy, Ruzena and Aloimonos, Yiannis and Tsotsos, John K},
  year = {2018},
  journal = {Autonomous Robots},
  volume = {42},
  number = {2},
  pages = {177--196}
}

@incollection{floyd1993assigning,
  title = {Assigning Meanings to Programs},
  booktitle = {Program Verification: {{Fundamental}} Issues in Computer Science},
  author = {Floyd, Robert W},
  year = {1993},
  pages = {65--81}
}

@phdthesis{he2025mathematical,
  type = {B.{{S}}. {{Thesis}}},
  title = {Mathematical Theory and Algorithms for Scence Semantics in Robotics},
  author = {He, Siming},
  year = {2025},
  school = {University of Pennsylvania}
}

@book{kirby2019invitation,
  title = {An Invitation to Model Theory},
  author = {Kirby, Jonathan},
  year = {2019},
  publisher={Cambridge University Press}
}


%% file: references.bib
@article{tao2025halo,
  title={HALO: High-Altitude Language-Conditioned Monocular Aerial Exploration and Navigation},
  author={Tao, Yuezhan and Ong, Dexter and Cladera, Fernando and Hughes, Jason and others},
  journal={IEEE RA-L},
  year={2026}
}

@inproceedings{strong2025next,
  title={{Next best sense: Guiding vision and touch with FisherRF for 3D gaussian splatting}},
  author={Strong, Matthew and Lei, Boshu and Swann, Aiden and Jiang, Wen and others},
  booktitle={ICRA},
  year={2025}
}

@inproceedings{Jiang2023FisherRF,
      title={FisherRF: Active View Selection and Uncertainty Quantification for Radiance Fields using Fisher Information},
      author={Wen Jiang and Boshu Lei and Kostas Daniilidis},
      booktitle={ICCV},
      year={2024}
  }

@inproceedings{tao2024learning,
  title={Learning to explore indoor environments using autonomous micro aerial vehicles},
  author={Tao, Yuezhan and Iceland, Eran and Li, Beiming and Zwecher, Elchanan and others},
  booktitle={ICRA},
  year={2024}
}

@article{maggio2024clio,
  title={Clio: Real-time task-driven open-set 3d scene graphs},
  author={Maggio, Dominic and Chang, Yun and Hughes, Nathan and Trang, Matthew and Griffith, Dan and Dougherty, Carlyn and Cristofalo, Eric and Schmid, Lukas and Carlone, Luca},
  journal={IEEE RA-L},
  year={2024}
}

@article{wang2025yoloe,
  title={Yoloe: Real-time seeing anything},
  author={Wang, Ao and Liu, Lihao and Chen, Hui and Lin, Zijia and Han, Jungong and Ding, Guiguang},
  journal={arXiv:2503.07465},
  year={2025}
}

@article{zhao2023fast,
  title={Fast segment anything},
  author={Zhao, Xu and Ding, Wenchao and An, Yongqi and Du, Yinglong and Yu, Tao and Li, Min and Tang, Ming and Wang, Jinqiao},
  journal={arXiv:2306.12156},
  year={2023}
}

@incollection{lorensen1998marching,
  title={Marching cubes: A high resolution 3D surface construction algorithm},
  author={Lorensen, William E and Cline, Harvey E},
  booktitle={Seminal graphics},
  year={1998}
}

@article{tourani2025vs,
  title={{vs-graphs: Integrating visual slam and situational graphs through multi-level scene understanding}},
  author={Tourani, Ali and Ejaz, Saad and Bavle, Hriday and others},
  journal={arXiv:2503.01783},
  year={2025}
}

@inproceedings{hu2023yoso,
  title={You only segment once: Towards real-time panoptic segmentation},
  author={Hu, Jie and Huang, Linyan and Ren, Tianhe and Zhang, Shengchuan and others},
  booktitle={CVPR},
  year={2023}
}

@article{fischler1981random,
  title={Random sample consensus: a paradigm for model fitting with applications to image analysis and automated cartography},
  author={Fischler, Martin A and Bolles, Robert C},
  journal={CACM},
  year={1981},
}

@inproceedings{siming2024active,
  title={Active perception using neural radiance fields},
  author={Siming, H and Hsu, Christopher D and Ong, Dexter and Shao, Yifei Simon and Chaudhari, Pratik},
  booktitle={ACC},
  year={2024}
}

@article{mildenhall2021nerf,
  title={{NERF: Representing scenes as neural radiance fields for view synthesis}},
  author={Mildenhall, Ben and Srinivasan, Pratul P and Tancik, Matthew and Barron, Jonathan T and Ramamoorthi, Ravi and Ng, Ren},
  journal={CACM},
  volume={65},
  number={1},
  pages={99--106},
  year={2021},
}

@inproceedings{ramakrishnan2021hm3d,
  title={{Habitat-Matterport 3D Dataset} ({HM}3D): 1000 Large-scale 3D Environments for Embodied {AI}},
  author={Santhosh Kumar Ramakrishnan and Aaron Gokaslan and Erik Wijmans and others},
  booktitle={NeurIPS Datasets and Benchmarks Track},
  year={2021}
}

@inproceedings{savva2019habitat,
  title={{Habitat: A platform for embodied AI research}},
  author={Savva, Manolis and Kadian, Abhishek and Maksymets, Oleksandr and others},
  booktitle={ICCV},
  year={2019}
}

@article{asgharivaskasi2023semantic,
  title={{Semantic octree mapping and Shannon mutual information computation for robot exploration}},
  author={Asgharivaskasi, Arash and Atanasov, Nikolay},
  journal={IEEE T-RO},
  year={2023},
}

@inproceedings{yamauchi1997frontier,
  title={{A frontier-based approach for autonomous exploration}},
  author={Yamauchi, Brian},
  booktitle={CIRA},
  year={1997}
}

@inproceedings{he2025active,
  title={{An Active Perception Game for Robust Information Gathering}},
  author={He, Siming and Tao, Yuezhan and Spasojevic, Igor and Kumar, Vijay and Chaudhari, Pratik},
  booktitle={ICRA},
  year={2025}
}

@inproceedings{bircher2016receding,
  title={{Receding horizon "next-best-view" planner for 3D exploration}},
  author={Bircher, Andreas and Kamel, Mina and Alexis, Kostas and Oleynikova, Helen and Siegwart, Roland},
  booktitle={ICRA},
  year={2016}
}

@inproceedings{hughes2022hydra,
  title={{Hydra: A Real-time Spatial Perception System for {3D} Scene Graph Construction and Optimization}},
  author={Nathan Hughes, Yun Chang and Luca Carlone},
  booktitle={RSS},
  year={2022},
}

@inproceedings{charrow2015information,
  title={{Information-theoretic mapping using Cauchy-Schwarz quadratic mutual information}},
  author={Charrow, Benjamin and Liu, Sikang and Kumar, Vijay and Michael, Nathan},
  booktitle={ICRA},
  year={2015}
}

@inproceedings{oleynikova2017voxblox,
  title={{Voxblox: Incremental 3D Euclidean signed distance fields for on-board MAV planning}},
  author={Oleynikova, Helen and Taylor, Zachary and Fehr, Marius and Siegwart, Roland and Nieto, Juan},
  booktitle={IROS},
  year={2017}
}

@article{kerbl20233d,
  title={{3D Gaussian splatting for real-time radiance field rendering}},
  author={Kerbl, Bernhard and Kopanas, Georgios and Leimk{\"u}hler, Thomas and Drettakis, George},
  journal={ACM Trans. Graph.},
  year={2023}
}

@article{chaplot2020object,
  title={Object goal navigation using goal-oriented semantic exploration},
  author={Chaplot, Devendra Singh and Gandhi, Dhiraj Prakashchand and Gupta, Abhinav and Salakhutdinov, Russ R},
  journal={NeurIPS},
  year={2020}
}

@inproceedings{semanticnerf,
  title={In-place scene labelling and understanding with implicit scene representation},
  author={Zhi, Shuaifeng and Laidlow, Tristan and Leutenegger, Stefan and Davison, Andrew J},
  booktitle={ICCV},
  year={2021}
}

@article{chen2023not,
  title={How to not train your dragon: Training-free embodied object goal navigation with semantic frontiers},
  author={Chen, Junting and Li, Guohao and Kumar, Suryansh and others},
  journal={RSS},
  year={2023}
}

@inproceedings{radford2021learning,
  title={Learning transferable visual models from natural language supervision},
  author={Radford, Alec and Kim, Jong Wook and Hallacy, Chris and others},
  booktitle={ICML},
  year={2021}
}

@article{comanici2025gemini,
  title={Gemini 2.5: Pushing the frontier with advanced reasoning, multimodality, long context, and next generation agentic capabilities},
  author={Comanici, Gheorghe and Bieber, Eric and Schaekermann, Mike and others},
  journal={arXiv:2507.06261},
  year={2025}
}

@inproceedings{yokoyama2024vlfm,
  title={{VLFM: Vision-language frontier maps for zero-shot semantic navigation}},
  author={Yokoyama, Naoki and Ha, Sehoon and Batra, Dhruv and others},
  booktitle={ICRA},
  year={2024}
}

@article{malladi2025robust,
  title={A Robust Approach for LiDAR-Inertial Odometry Without Sensor-Specific Modeling},
  author={Malladi, Meher VR and Guadagnino, Tiziano and Lobefaro, Luca and Stachniss, Cyrill},
  journal={arXiv:2509.06593},
  year={2025}
}

@inproceedings{millane2024nvblox,
  title={nvblox: Gpu-accelerated incremental signed distance field mapping},
  author={Millane, Alexander and Oleynikova, Helen and Wirbel, Emilie and Steiner, Remo and Ramasamy, Vikram and Tingdahl, David and Siegwart, Roland},
  booktitle={ICRA},
  year={2024},
}

@inproceedings{dai2024optimal,
  title={Optimal scene graph planning with large language model guidance},
  author={Dai, Zhirui and Asgharivaskasi, Arash and Duong, Thai and others},
  booktitle={ICRA},
  year={2024}
}

@inproceedings{song2017semantic,
  title={Semantic scene completion from a single depth image},
  author={Song, Shuran and Yu, Fisher and Zeng, Andy and others},
  booktitle={CVPR},
  year={2017}
}

@inproceedings{cao2022monoscene,
  title={Monoscene: Monocular 3d semantic scene completion},
  author={Cao, Anh-Quan and De Charette, Raoul},
  booktitle={CVPR},
  year={2022}
}

@inproceedings{wan2018representation,
  title={Representation Learning for Scene Graph Completion via Jointly Structural and Visual Embedding.},
  author={Wan, Hai and Luo, Yonghao and Peng, Bo and Zheng, Wei-Shi},
  booktitle={IJCAI},
  year={2018}
}

@inproceedings{garg2021unconditional,
  title={Unconditional scene graph generation},
  author={Garg, Sarthak and Dhamo, Helisa and Farshad, Azade and Musatian, Sabrina and Navab, Nassir and Tombari, Federico},
  booktitle={ICCV},
  year={2021}
}

@inproceedings{anderson2018vision,
  title={Vision-and-language navigation: Interpreting visually-grounded navigation instructions in real environments},
  author={Anderson, Peter and Wu, Qi and Teney, Damien and Bruce, Jake and Johnson, Mark and S{\"u}nderhauf, Niko and Reid, Ian and Gould, Stephen and Van Den Hengel, Anton},
  booktitle={CVPR},
  year={2018}
}

@article{zhang2024uni,
  title={Uni-navid: A video-based vision-language-action model for unifying embodied navigation tasks},
  author={Zhang, Jiazhao and Wang, Kunyu and Wang, Shaoan and Li, Minghan and Liu, Haoran and Wei, Songlin and Wang, Zhongyuan and Zhang, Zhizheng and Wang, He},
  journal={arXiv:2412.06224},
  year={2024}
}

@article{kim2024openvla,
  title={Openvla: An open-source vision-language-action model},
  author={Kim, Moo Jin and Pertsch, Karl and Karamcheti, Siddharth and Xiao, Ted and Balakrishna, Ashwin and Nair, Suraj and Rafailov, Rafael and Foster, Ethan and Lam, Grace and Sanketi, Pannag and others},
  journal={arXiv:2406.09246},
  year={2024}
}
